\documentclass[sigconf]{aamas}  

\usepackage{booktabs}

\usepackage{graphicx}

\usepackage{flushend}
\setcopyright{ifaamas}  
\acmDOI{doi}  
\acmISBN{}  
\acmConference[AAMAS'20]{Proc.\@ of the 19th International Conference on Autonomous Agents and Multiagent Systems (AAMAS 2020), B.~An, N.~Yorke-Smith, A.~El~Fallah~Seghrouchni, G.~Sukthankar (eds.)}{May 2020}{Auckland, New Zealand}  
\acmYear{2020}  
\copyrightyear{2020}  
\acmPrice{}  

\usepackage{xcolor}
\usepackage{amssymb}
\usepackage{amsfonts}
\usepackage{graphicx}
\usepackage{mathtools}
\usepackage{amsmath}
\usepackage{gensymb}
\usepackage{float}
\usepackage{multirow}
\usepackage{makecell}
\usepackage[utf8]{inputenc}
\usepackage[english]{babel}

\usepackage{amsthm}

\usepackage{subcaption}

\newcommand{\gray}[1]{\textcolor{gray}{#1}}

\newcommand\Tstrut{\rule{0pt}{2.6ex}}         

\begin{document}

\title{Realtime Collision Avoidance for Mobile Robots in Dense Crowds using Implicit Multi-sensor Fusion and Deep Reinforcement Learning}  


\author{Jing Liang*, Utsav Patel*, Adarsh Jagan Sathyamoorthy, and Dinesh Manocha.}
\authornote{Authors contributed equally.}
\authornote{Video: https://youtu.be/N83Mg9oW-0c}

\begin{abstract}  
We present a novel learning-based collision avoidance algorithm, CrowdSteer, for mobile robots operating in dense and crowded environments. Our approach is end-to-end and uses multiple perception sensors such as a 2-D lidar along with a depth camera to sense surrounding dynamic agents and compute collision-free velocities. Our training approach is based on the sim-to-real paradigm and uses high fidelity 3-D simulations of pedestrians and the environment to train a policy using Proximal Policy Optimization (PPO). We show that our learned navigation model is directly transferable to previously unseen virtual and dense real-world environments. We have integrated our algorithm with differential drive robots and evaluated its performance in narrow scenarios such dense crowds, narrow corridors, T-junctions, L-junctions, etc. In practice, our approach can perform real-time collision avoidance and generate smooth trajectories in such complex scenarios. We also compare the performance with prior methods based on metrics such as trajectory length, mean time to goal, success rate, and smoothness and observe considerable improvement.
\end{abstract}


\maketitle


\linespread{0.93}
\section{Introduction}
Mobile robots are frequently deployed in indoor and outdoor environments such as hospitals, hotels, malls, airports, warehouses, sidewalks, etc. These robots are used for surveillance, inspection, delivery, and cleaning, or as social robots. Such applications need to be able to smoothly and reliably navigate in these scenarios by avoiding collisions with obstacles, including dynamic agents or pedestrians.

Some earlier work on mobile robot navigation was limited to open spaces or simple environments with static obstacles. Over the last decade, there has been considerable work on collision-free navigation among pedestrians using visual sensors like lidars or cameras \cite{FoxThurn,WB1,End2End,JHow1,7989182,Unfrozen}. 
However, many challenges arise when such robots are used in dense or cluttered environments and need to move at speeds that are close to that of  human pedestrians (~1.3 meters/sec). A high crowd density corresponds to 1-3 (or more) pedestrians per square meter. In these scenarios, the pedestrian trajectories are typically not smooth and may change suddenly. Moreover, it is difficult to predict their trajectories due to occlusion or non-smooth motion. Many sensor-based navigation algorithms either tend to stall the robot's motion or are unable to avoid collisions with the pedestrians.


A recent trend is to use learning methods for sensor-based robot navigation in crowds. These include techniques based on end-to-end deep learning~\cite{End2End,7989182}, generative adversarial imitation learning~\cite{WB1}, and deep reinforcement learning \cite{JiaPan1,JiaPan2}. Most of these methods use one or more perception sensors like an RGB, or RGB-D camera, or a 2-D lidar.  


\begin{figure}[t]
\begin{tabular}{ll}
\includegraphics[width=.45\columnwidth]{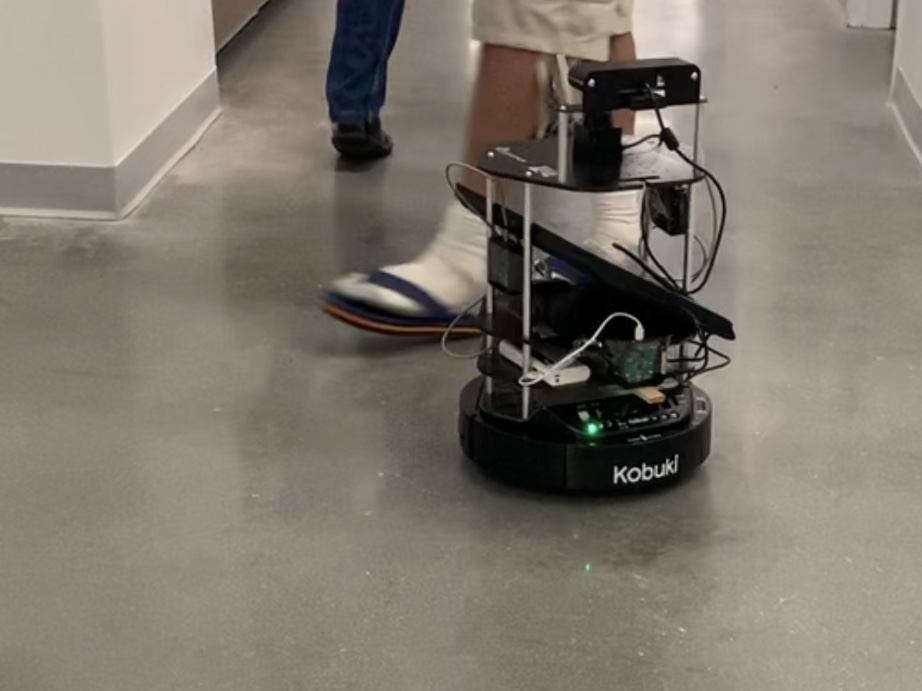}
&
\includegraphics[width=.45\columnwidth]{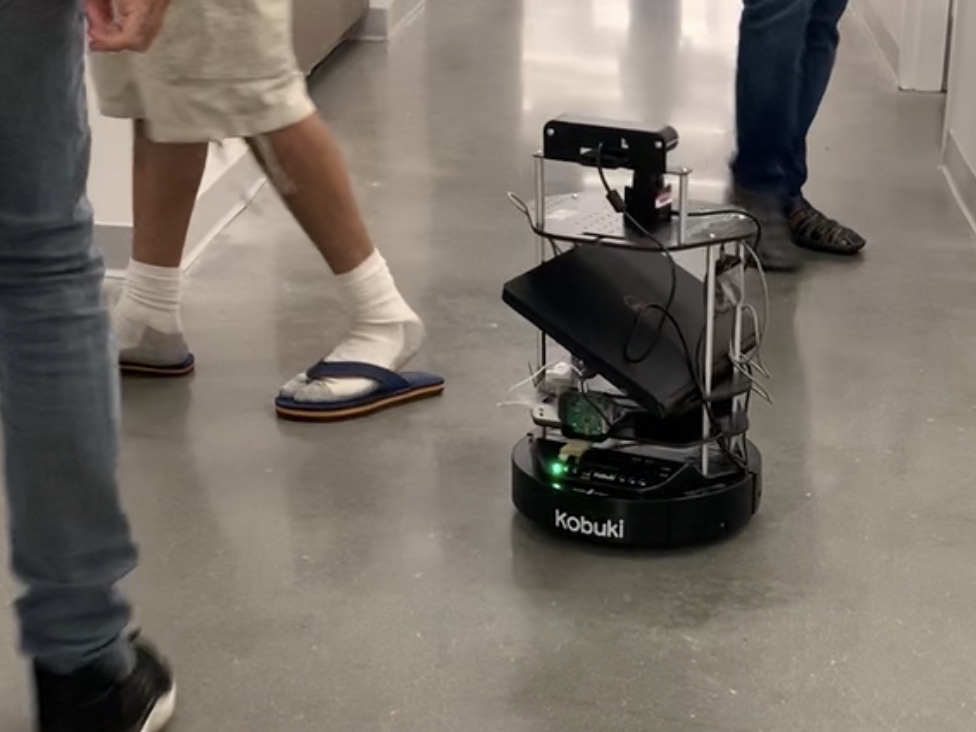}
\end{tabular}
\begin{tabular}{ll}
\includegraphics[width=.45\columnwidth]{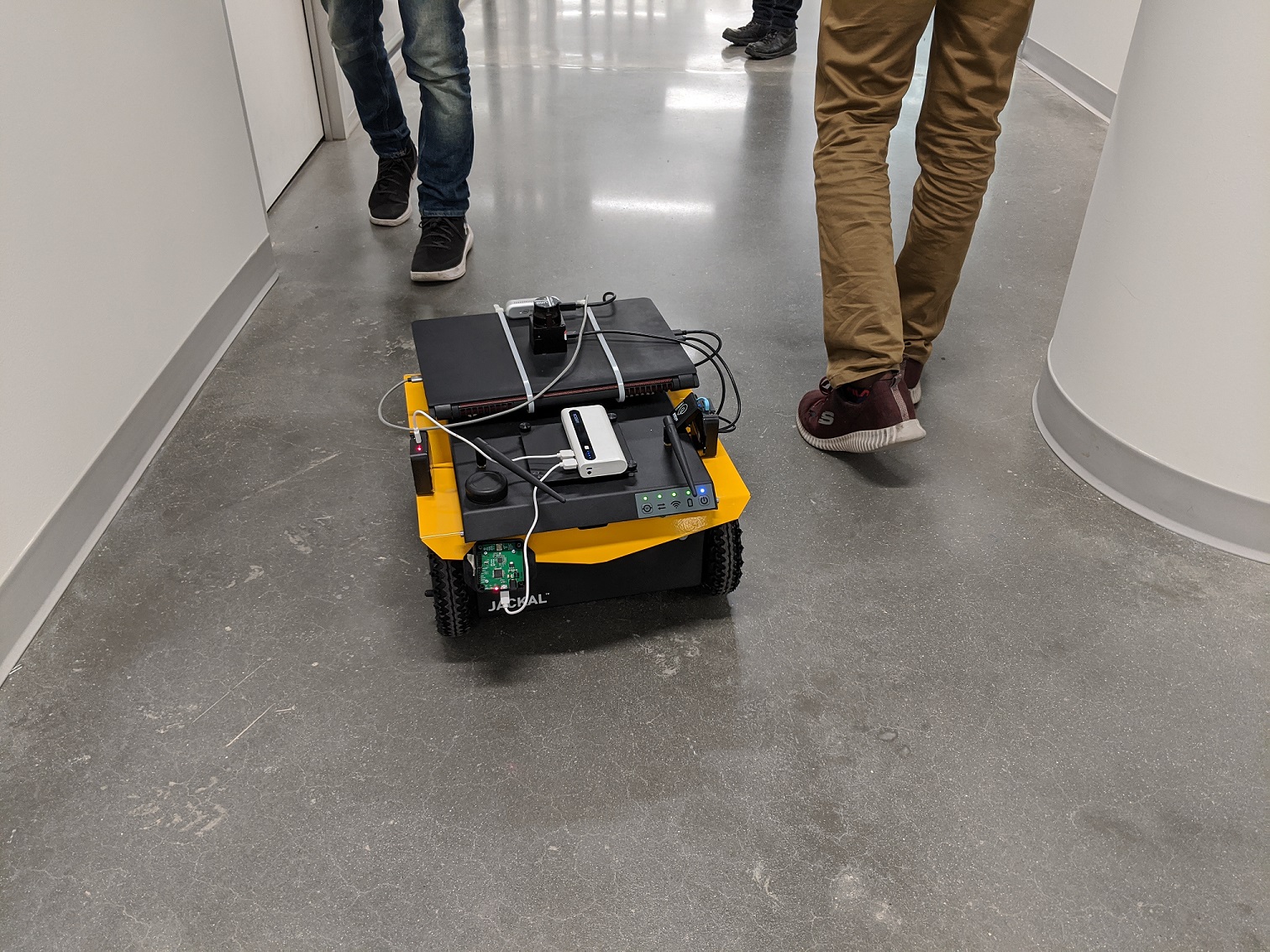}
&
\includegraphics[width=.45\columnwidth]{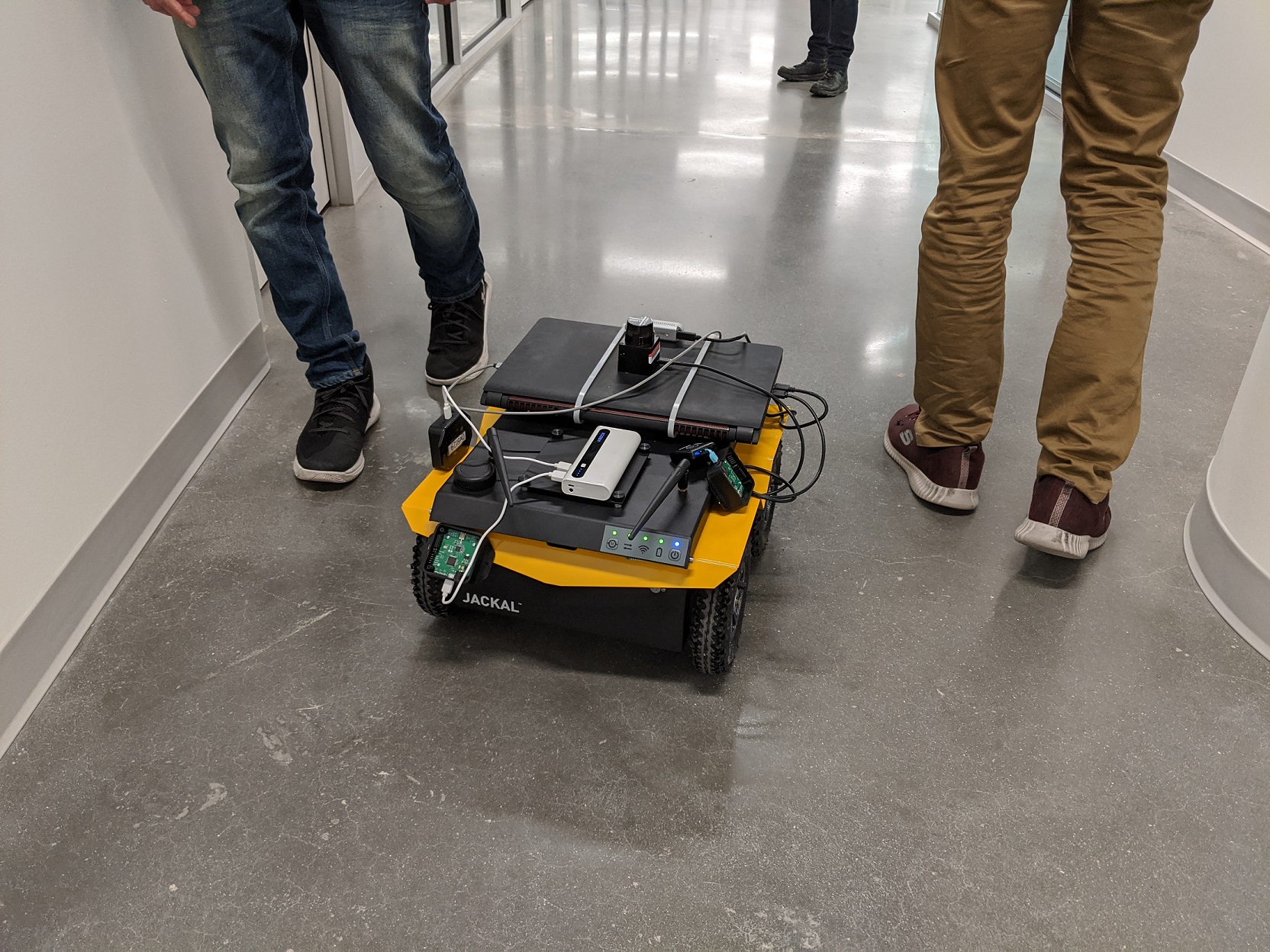}
\end{tabular}
\caption{A Turtlebot and Jackal robot using CrowdSteer to navigate in scenarios with pedestrians in a narrow corridor and areas with high occlusion. Our method uses data from multiple sensors such as a 2-D lidar and a depth camera to generate smooth collision avoidance maneuvers. We compare with methods such as DWA\cite{DWA} and Fan et al\cite{JiaPan1}.}
\label{fig:confusionmatrix}
\vspace{-10pt}
\end{figure}

In practice, using a single robot sensor may not work well in dense or cluttered scenarios. Moreover,  this sensor choice affects the efficiency of the collision avoidance scheme in terms of  reaction time or the optimality of the trajectory. Some of these methods work well in static environments, but fail in scenarios with even a few dynamic obstacles~\cite{WB1}. Algorithms that use  a lidar may perform well in dense scenarios but lack the ability to detect thin obstacles such as poles or differentiate between animate and inanimate obstacles. They also suffer from the freezing robot problem and exhibit oscillatory behaviors in dense situations. Other algorithms~\cite{JHow1,JHow2} use either 2-D or 3-D lidars along with several RGB cameras to detect obstacles and predict pedestrian motions. They exhibit good performance in moderately dense crowds but may not work well with high density crowds, scenes with occlusions and are susceptible to perception errors. 


{\bf Main Results:} We present CrowdSteer, a novel learning-based local navigation method that uses hybrid sensing for densely crowded scenarios.  We formulate the navigation among pedestrians as a POMDP problem and solve it using deep reinforcement learning. Our approach uses a combination of perception sensors such as a 2-D lidar along with an depth (RGB-D) camera to sense obstacle features and help the policy implicitly learn different kinds of interactions with the obstacles. We use the Proximal Policy Optimization (PPO) algorithm to train the collision avoidance policy and reduce the sim-to-reality gap by using high-fidelity complex simulations of environments with pedestrians. Some of the novel components of our approach include:
\begin{itemize}
\item A new end-to-end learning method that fuses inputs from multiple perception sensors for implicitly characterizing the robot's interactions with obstacles and pedestrians. This results in reliably handling complex, high-density scenarios for navigation.

\item A trained model which generates smoother collision avoidance trajectories and is resilient to perception noises and occluded obstacles, as compared to previous methods.

\item Custom made complex high-fidelity training and testing scenarios of indoor environments with pedestrians for training and testing Deep Reinforcement Learning (DRL) models. We show that using such simulations eases sim-to-real transfer and leads to better generalization of the trained policy.
\end{itemize}

We implemented and evaluated our algorithm on a Turtlebot robot and a ClearPath Jackal robot with a Hokuyo 2-D lidar and an Orbbec Astra camera in indoor environments such as corridors, L-junctions, and T-junctions with varying pedestrian density (as high as 1-2 humans/ $m^2$). We also compare CrowdSteer's performance with prior traditional methods such as Dynamic Window Approach (DWA)~\cite{DWA} and a state-of-the-art learning-based crowd navigation algorithm~\cite{JiaPan1}. We observe that our approach surpasses these methods in terms of success rates, and shows a reduction of up to 68.16\%  in time to goal, and 6.12\% reduction in trajectory length when compared to the current state of the art Fan et al. ~\cite{JiaPan1}.


\section{Related Work}
In this section, we give a brief overview of the prior work on traditional and learning-based navigation algorithms.

\subsection{Navigation in Dynamic Scenes}
There is extensive work on collision avoidance in dynamic scenes for robots. These include techniques based on potential-field methods~\cite{PFM1}, social-forces~\cite{helbing1995social}, velocity obstacles~\cite{ORCA,NH-ORCA}, etc. These methods have been used in simulated environments and can scale to a large number of agents.  In terms of real-world scenarios, these  methods require accurate sensing of obstacles' positions and velocities and parameter tuning that is scenario-dependent. 
These requirements make it difficult to directly apply them for navigation in dense crowds. At the same time, these methods have been used to generate training trajectories for initializing some learning-based methods.
The Dynamic Window Approach (DWA)~\cite{DWA} is another widely used method which calculates reachable dynamically-constrained velocities for collision avoidance within a short time interval. However, it does not scale well to large numbers of dynamic obstacles.

\subsection{Sensor-based Navigation among Pedestrians}
Sensor-based navigation algorithms are widely used to navigate a robot among pedestrians~\cite{FoxThurn}. In \cite{IMM}, data from a radar and far infrared (IR) camera were synchronized and fused to track obstacles for collision avoidance. A significant problem in navigating among pedestrians is modeling their unknown intentions. \cite{Thompson} used the observed motions of humans to generate a motion probability grid to model pedestrian intentions. Some methods \cite{V-obstacles} learned obstacle motions from trajectories in a captured video or used laser scan data and Hidden Markov Models to estimate human trajectories \cite{Thrun}.   

Other techniques use a Partially Observable Markov Decision Process (POMDP) to model the uncertainties in the intentions of pedestrians. A POMDP-based planner was presented in \cite{POMDP} to estimate the pedestrians goals for autonomous driving which was later augmented with an ORCA-based pedestrian motion model~\cite{PORCA}. The resulting POMDP planner runs in near real-time and is able to choose actions for the robot. Our  approach is complimentary to these methods in that we model the navigation problem as a POMDP. 

\subsection{Learning-based Collision Avoidance}
There is considerable work on using different learning methods for collision avoidance and navigation in dense environments.

\subsubsection{\textbf{Using single perception sensor:}} 
Several works have used data from a single perception sensor to train collision avoidance behaviors in a robot. A map-less navigation method using expert demonstrations in simulation was trained with a single 2-D lidar for static environments in ~\cite{CesarCadena}. 
In \cite{WB1} a GAIL (Generative Adversarial Imitation Learning) strategy was trained using raw data from a depth camera over a pre-trained behavioral cloning policy to generate socially acceptable navigation through a crowd. However, the robot's navigation is limited by the depth camera's field of view (FOV) and works well only in sparse crowds and near-static environments. 

An end-to-end visuomotor navigation system using CNNs that are trained directly with RGB images was developed in \cite{End2End}. Similarly, \cite{D3QN} used a deep double-Q network (D3QN) to predict depth information from RGB images and used it for static obstacle avoidance in cluttered environments. \cite{Target_driven} improved the generalization capability of deep reinforcement learning by including a visual goal in the policy of their actor-critic models. While these methods perform well for mostly static scenarios, they may not work well with dense crowds.

A decentralized sensor-level collision avoidance method that was trained with multi-robot Proximal Policy Optimization (PPO) \cite{PPO} using a 2-D lidar in \cite{JiaPan1}. 
This approach was extended in \cite{JiaPan2} based on a hybrid control architecture, which switched between different policies based on the density of the obstacles in the environment. This approach works well in open spaces, but exhibits oscillatory and jerky motions in dense scenarios since the 2-D lidar only senses the proximity data and fails to sense more complex interactions. 

\subsubsection{\textbf{Using multiple sensors:}} 
\cite{JHow1} presents a decentralized agent-level collision avoidance method by utilizing a trained value network that models the cooperative behaviors in multi-agent systems.
An LSTM-based strategy that uses observations of arbitrary numbers of neighboring agents during the training phase and makes no assumptions about obstacles' behavior rules is described in \cite{JHow2}. Other methods~\cite{Alahi} explicitly model robot-human interactions in a crowd for robot navigation. These algorithms use a 2-D or 3-D lidar along with several RGB cameras for pedestrian classification and obstacle detection. However, these methods use a time parameter ($\Delta t$) for which obstacle motions are assumed to be linear. The value of $\Delta t$ is important for their training to converge and their performances are susceptible to perception errors. In contrast, our training process is more robust and makes no such assumptions.

\cite{JHow-uncertainty} presents an uncertainty-aware reinforcement learning method for collision avoidance that identifies novel scenarios and performs careful actions around the pedestrians. A method to solve both the freezing robot problem and loss of localization simultaneously by training an actor-critic model to learn localization recovery points in the environment is shown in \cite{Unfrozen}. This approach was also based on a single sensor and susceptible to jerky/oscillatory motion.

\section{Overview}
In this section, we introduce our multi-sensor based navigation problem and give an overview of our approach based on deep reinforcement learning. Unlike prior methods, our goal is to simultaneously use multiple sensors that sense the proximity data and various interactions and behaviors of dynamic agents and pedestrians and generate smooth trajectories.

      
      
      
      
      
      
      
      
      
      
   
\subsection{Robot Navigation}
We assume that the dynamics of the robot we train is bounded by non-holonomic constraints~\cite{NH-ORCA}. The robot at any point in time knows its environment only to the extent of its sensor observations, and no global knowledge of the state of the environment may be available. In addition, in accordance with real-world scenarios, the robot may not be able to access other hidden parameters of pedestrians and other dynamic agents, such as goals and states. At each time step $t$, the  robot  has access to an observation vector $o^t$ which it uses to compute a collision-free action that drives it towards its goal from the current position by avoiding collisions with the static and dynamic obstacles. We assume that it is the sole responsibility of the robot to avoid collisions with pedestrians.

\subsection{Multiple Sensors}
Our approach is designed to exploit multiple sensors simultaneously. These include a lidar to measure the proximity to different obstacles and pedestrians. However, the lidar's raw data does not provide sufficient information to differentiate between animate and inanimate obstacles or sudden changes in the orientation of obstacles. As a result, it is difficult to infer whether the obstacles  are moving towards or away from the robot. Therefore, we also use RGB or depth cameras for such observations. Furthermore, these cameras are able to capture the interactions between the obstacles and pedestrians in the scene.

\subsubsection{\textbf{2-D lidar:}} Each scan/frame from a 2-D lidar consists of a list of distance values on the plane of sight of the lidar (See left scenario in Fig. \ref{fig:TrainingScenarios}). With its high accuracy, field of view (FOV), and low dimensional output data, the 2-D lidar allows us to detect clusters of closest points in the robot's surroundings.

\subsubsection{\textbf{Cameras:}} 
\begin{figure}[t]
      \centering
      \includegraphics[height=1.7in,width=3.2in]{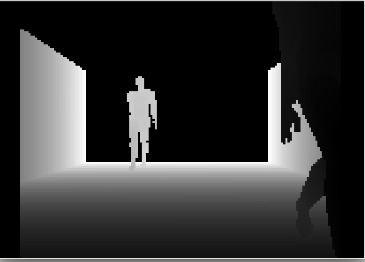}
      \caption {Contours of a pedestrian walking at a distance, and surrounding walls captured by the depth camera on our robot. Features such as the pedestrian's pose and motion can be extracted from such frames. Our CrowdSteer algorithm exploits these features for collision avoidance and generating smooth trajectories.}
      \label{fig:contour}
   \end{figure}

Depth images possess an additional dimension over and above a 2-D lidar. Therefore, features such as obstacle contours and changes in obstacles' poses are more prominently recorded even in low resolution images (Fig \ref{fig:contour}). This facilitates feature extraction to differentiate between moving and non-moving objects. For instance, a pedestrian changing direction away from the robot's trajectory would result in a contour with a lower area in the depth image. If we consider several consecutive frames from the camera, the approximate positions, orientations and velocities of all obstacles in the frame can be extracted. The same principles apply to RGB or grayscale images from an RGB camera. RGB cameras, in general, have a higher FOV than depth cameras and have an infinite sensing range similar to the human eye. Therefore, features corresponding to the obstacles can be captured in a single frame even when they are far from the robot.

The lidar and the camera sensors provide complimentary information about the environment. Using multiple consecutive frames of this combined information, helps our method learn that a change in features in the sensor data leads to a change in interaction with the obstacles and produces actions to accommodate it. In addition, a sensor's limited FOV can be overcome if another sensor with a high FOV is used in tandem with it. Using such combination of sensors is highly useful in indoor scenarios with a lot of occlusion. In our work, we assume that the data from the 2-D lidar and the camera are synchronized.

\subsection{Problem Formulation}
We formulate the navigation among pedestrians as a POMDP, which is solved using deep reinforcement learning. Formally, a POMDP is modeled as a 6-tuple (S, A, P, R, $\Omega$, O) \cite{JiaPan2}, where the symbols represent the state space, action space, state-transition model, reward function, observation space and observation probability distribution given the system state, respectively. As stated earlier, our robot has access only to the observations, which can be sampled from the system's state space. Next, we describe our observation and action spaces.   

Using data from the 2-D lidar along with a camera makes our observation space high-dimensional. In addition, since the individual sensor streams have different dimensions (1-D for lidar and 2-D for images), they cannot be processed together. Therefore, we split the robot's observation vector into four components, $\textbf{o}^t = [ \textbf{o}^t_{lid}, \textbf{o}^t_{cam}, \textbf{o}^t_g, \textbf{o}^t_v ]$, where $\textbf{o}^t_{lid}$ denotes raw noisy 2-D lidar measurements, $\textbf{o}^t_{cam}$ denotes the raw image data from either a depth or an RGB camera, $\textbf{o}^t_g$ refers to the relative goal location with respect to the robot, and $\textbf{o}^t_v$ denotes the current velocity of the robot. $\textbf{o}^t_{lid}$ is mathematically represented as:
\begin{equation}
    \textbf{o}^t_{lid} = \{l \in \mathbb{R}^{512} : 0 < l_i < 4\}
    \label{lidar}
\end{equation}

Where \textit{l} represents a list of proximity values and $l_i$ denotes the $i^{th}$ element of the list. $\textbf{o}^t_{cam}$ can be mathematically denoted as:
\begin{equation}
    \textbf{o}^t_{cam} = \{C \in \mathbb{R}^{150 \times 120} : 1.4 < C_{ij} < 5\}
    \label{cam}
\end{equation}

The {\em action space} of the robot is composed of its linear and angular velocities $\textbf{a}^t = [v^t, \omega^t]$. The objective of the navigation algorithm is to select an action $\textbf{a}^t$ at each time instance, sampled from a trained policy $\pi_{\theta}$ as: 
\begin{equation}
    \textbf{a}^t \sim \pi_{\theta}(\textbf{a}^t | \textbf{o}^t).
\end{equation}

\noindent This action drives the robot towards its goal while avoiding collisions with pedestrians and static obstacles, until a new observation $\textbf{o}^{t+1}$ is measured. We use the minimization of the mean arrival time of the robot to its goal position as the objective function to optimize the policy $\pi_{\theta}$ as:
 \begin{equation}
     \underset{\pi_{\theta}}{\operatorname{argmin}}  \mathop{\mathbb{E}}[\frac{1}{N}\sum_{i = 1}^{N}t^g_i | \pi_{\theta}].
     \label{eqn1}
 \end{equation}
 
\subsection{Reinforcement Learning Training and PPO}
We use a policy gradient based \cite{sutton} reinforcement learning method called Proximal Policy Optimization (PPO)~\cite{PPO} to solve the optimization problem in Equation \ref{eqn1}. We adapt a policy gradient method since it directly models the strategy that generates actions, given the observations from the agents, and is more suitable for continuous action spaces such as ours. Compared with other policy gradient methods, PPO provides better stability during training by bounding the parameter ($\theta$) updates to a trust region, i.e., it ensures that the updated policy does not diverge from the previous policy. At each training episode, a robot in simulation collects a batch of observations until a time $T_{max}$ and the policy is then updated based on a loss function. PPO uses a surrogate loss function which is optimized using the Adam optimizer under the Kullback-Lieber(KL) divergence constraint which is given as:
\begin{align}
&\begin{aligned}
    L^{PPO}(\theta) &= \sum_{t=1}^{T_{max}} \frac{\pi_{\theta}(a^t_i | o^t_i)}{\pi_{old}(a^t_i | o^t_i)}\hat{A^t_i} - \beta KL[\pi_{old} | \pi_{\theta}] \\
    & + \xi max(0,KL[\pi_{old} | \pi_{\theta}] - 2KL_{target})^2
\end{aligned}
\end{align}
Where $\hat{A^t_i}$ is the advantage function, $\beta$ and $\xi$ are hyperparameters.  
The key issue is to design methods that take into account multiple sensor inputs and make sure that the training module converges fast and is able to handle all kind of observations.

\section{Our Approach: CrowdSteer}
In this section, we present our sensor-fusion based collision avoidance method that directly maps multiple sensor observations to a collision-free action. We describe the network that models our policy.

\subsection{Network Architecture}
\begin{figure}[t]
      \centering
      \includegraphics[height=2.0in,width=3.6in]{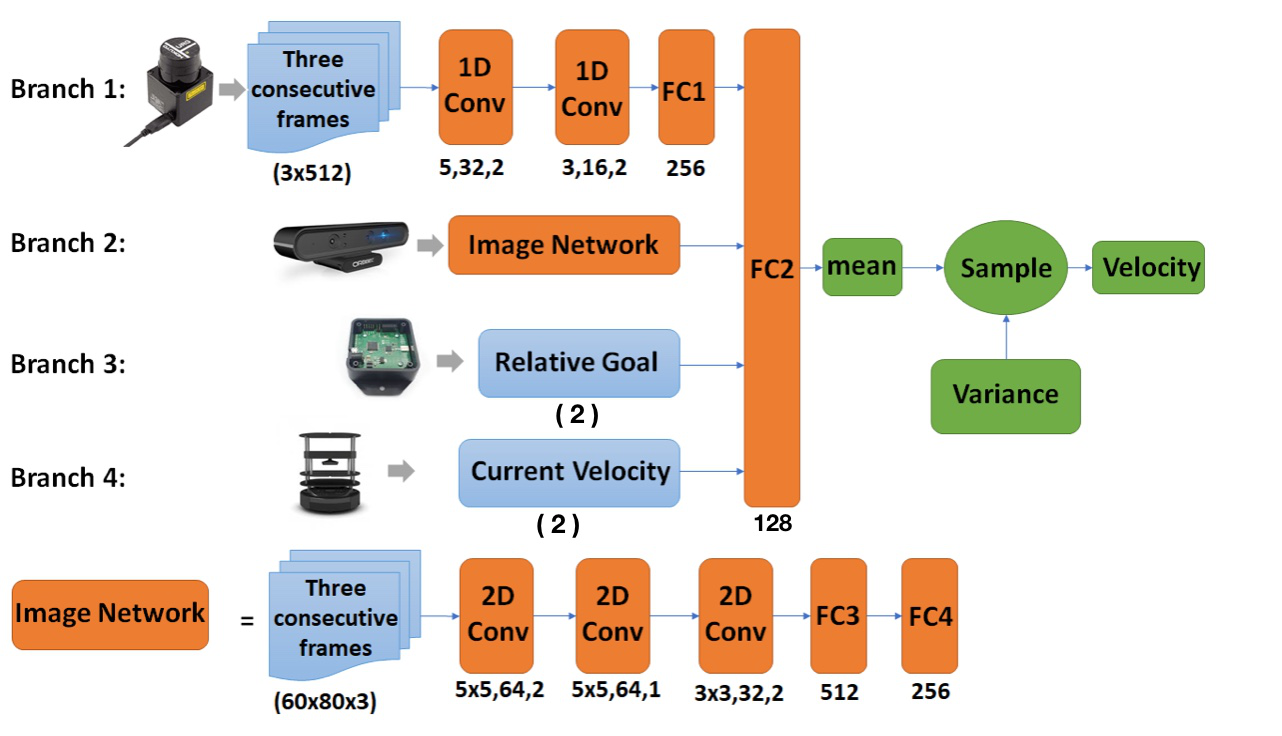}
      \caption {Architecture of our hybrid sensing network with four branches for different observations. The input layer is marked in blue, the hidden layers are marked in orange. Fully connected layers in the network are marked as FCn. The second branch extracts features from three consecutive image frames, which are fused with features extracted from three frames of the lidar in FC2 layer. The three values underneath each hidden layer denote the kernel size, number of filters, and stride length respectively.}
      \label{fig:Network}
   \end{figure}

Since the data from the lidar and camera have different dimensions (see Eqns \ref{lidar}, \ref{cam}), they cannot be processed together. Therefore, our network (Fig.\ref{fig:Network}) consists of four branches, each processing a single component of the observation $o^t$. 

Branches 1 and 2 consist of multiple 1-D and 2-D convolutional layers to process the lidar and image observations, respectively, as the convolutional layers exhibit  good performance in terms of extracting features from the input data \cite{imagenet}. Three consecutive lidar frames, each containing a list of proximity values are fed into two 1-D convolutional layers and a fully connected layer for processing. The first hidden layer has $32$ filters and stride length = 2. The second hidden layer has 16 filters, and stride length = 2. Kernels, which capture the relationship among nearby elements in an array, of size 5 and 3 are used in the convolutional layers respectively. 

Branch 2 is used to learn to detect motion from the three image frames, characterize the interaction based on obstacle motion and to move towards the free space detected in the frames. Three consecutive image frames are fed into a 2-D convolutional layer with 64 filters, kernel size = 5, and stride length = 2. The next layer uses 64 2-D convolutional filters, kernel size = 5 and stride length = 1. The output of this layer is passed on to 32 2-D convolutional layers of kernel size = 3, and stride length = 2. This is followed by two fully connected layers of sizes 512 and 256. Layer FC4 ensures that the output of branch 2 has the same dimensions as the output of branch 1. Layers FC1 and FC4 at the end of branches 1 and 2 also ensure that the data from the perception sensors have the most effect on the actions generated by the network. ReLU activation is applied to the outputs of all hidden layers in branches 1 and 2. 

The output of branches 1 and 2 and the goal and current velocity observations are then correlated together by the fully connected layer FC2 with 128 rectifier units. In the output layer, a sigmoid activation is used to restrict the robot's linear velocity between (0.0, 1.0) m/s and a tanh function restricts the angular velocity between (-0.4, 0.4) rad/s. The output velocity is sampled from a Gaussian distribution which uses the mean and log standard deviation which were updated during training. The training of the network is end-to-end and all parts of the network are trained simultaneously. 


\subsection{Reward Function}
We train the robot to reach its goal in the least possible time, while avoiding the obstacles. Therefore, the robot is rewarded for heading towards and reaching its goal, and penalized for moving too close or colliding with an obstacle. In addition, the robot is expected to avoid oscillatory velocities, follow smooth trajectories and reach intermediate waypoints before reaching the goal. 
Although the penalty for collision teaches the robot not to collide, it does not specifically result in the robot maintaining a safe distance from obstacles. This needs to be considered when the dynamic obstacles are pedestrians. Some previous algorithms~\cite{JiaPan1,JHow1} train their model for multi-agent collision avoidance, which results in the robot not maintaining a safe distance from pedestrians. The intermediate waypoints provide the robot with a sense of direction and guide it towards its goal. Formally, the total reward collected by a robot \textit{i} at time instant \textit{t} can be given as:
\begin{equation}
    \textit{r}_i^t = (r_g)^t_i \,+ (r_c)^t_i \,+ (r_{osc})^t_i \,+ (r_{safedist})^t_i
\end{equation}
where the reward for reaching the goal $(r_g)^t_i$ or an intermediate waypoint is given as:

\begin{equation}
    (r_g)^t_i =
    \begin{cases}
     r_{wp} \qquad \qquad \qquad \qquad if \, ||\textbf{p}_i^t - \textbf{p}_{wp}|| < 0.1,\\
     r_{goal} \qquad \qquad \qquad \qquad if \, ||\textbf{p}_i^t - \textbf{g}_i|| < 0.1,\\
     2.5(||p_i^{t-1} - g_i|| - ||p_i^t - g_i||) \qquad   otherwise.
    \end{cases}
\end{equation}
The collision penalty $(r_c)^t_i$ is given as: 
\begin{equation}
    (r_c)^t_i =
    \begin{cases}
     r_{collision} & if \, ||\textbf{p}_i^t - \textbf{p}_{obs}|| < 0.3,\\
     0  &  otherwise.
    \end{cases}
\end{equation}
The oscillatory behaviors (choosing sudden large angular velocities) are penalized as:
\begin{equation}
    (r_{osc})^t_i = -0.1 |\omega_i^t| \qquad \qquad  if \, |\omega_i^t| > 0.3.
    \label{oscillation}
\end{equation}
The penalty for moving too close to an obstacle is given by:
\begin{equation}
    (r_{safedist})^t_i = -0.1 ||R_{S_{max}} - R_{min}^t||.
\end{equation}
We set $r_{wp}$ = 10, $r_{goal}$ = 20, and $r_{collision}$ = -20 in our formulation. Critical behaviors such as collision avoidance and goal reaching have a higher priority in terms of the overall reward collected by the robot, while choosing smoother velocities and maintaining a safe distance from obstacles contribute to the reward with a slightly lower priority.

\subsection{Training Scenarios}
A major challenge in learning based methods is to close the sim-to-reality gap that arises when using a policy trained using simulated environments and sensor observations is evaluated in the real world. Additionally, when using multiple sensors, especially cameras, the simulation should contain as many real world features or characteristics as possible for the training to generalize well. One of our goals is to take advantage of the camera's ability to observe such complex features (Fig. \ref{fig:contour}). To address these issues simultaneously, we use high-fidelity 3-D environments that replicate real-world open spaces, and indoor scenarios with realistic moving pedestrians, and occluded environments.  

The policy training is carried out in multiple stages, starting with a low-complexity static scenario, to dynamic scenarios with pedestrians. The simple scenarios initialize the policy $\pi_{\theta}$ with capabilities such as static collision avoidance and goal-reaching, while dynamic collision avoidance capability is learned in complex scenarios with moving pedestrians. The different type of environments used in training are shown in (Fig.\ref{fig:TrainingScenarios}). These include:

\begin{figure*}[t]
\begin{tabular}{lll}
\includegraphics[width=.275\linewidth]{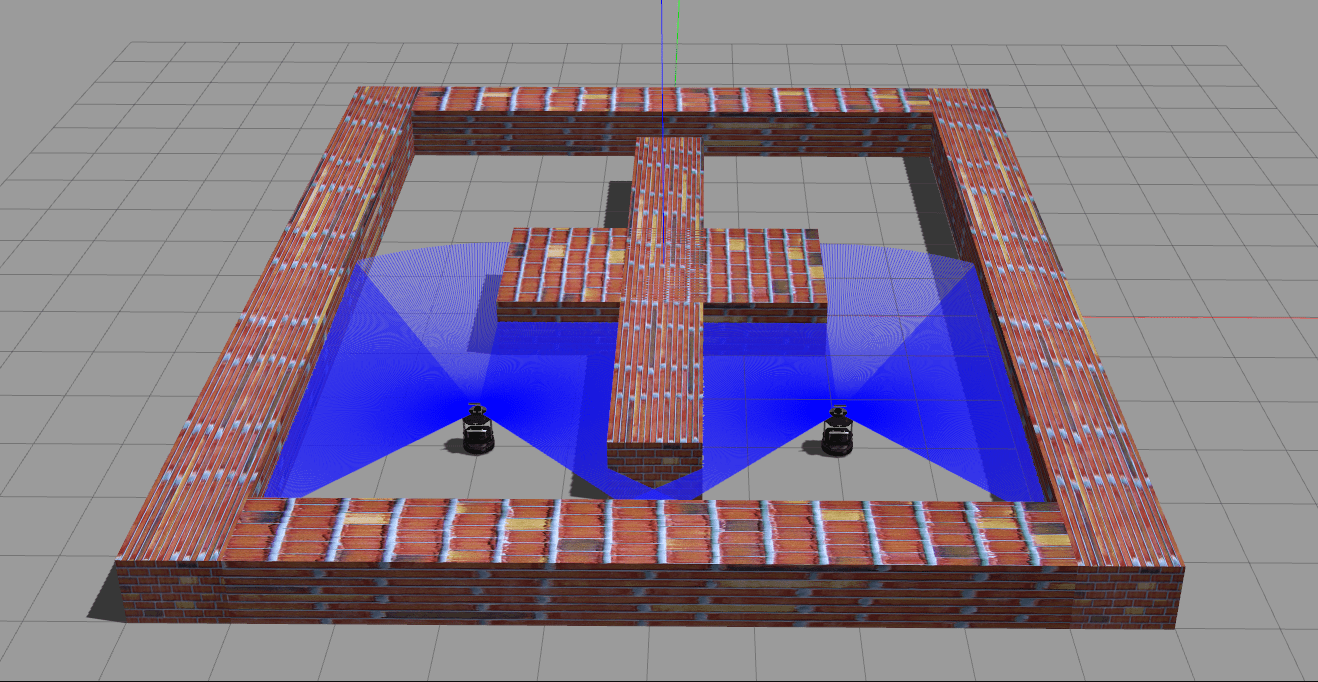}
&
\includegraphics[width=.25\linewidth, height=2.5cm]{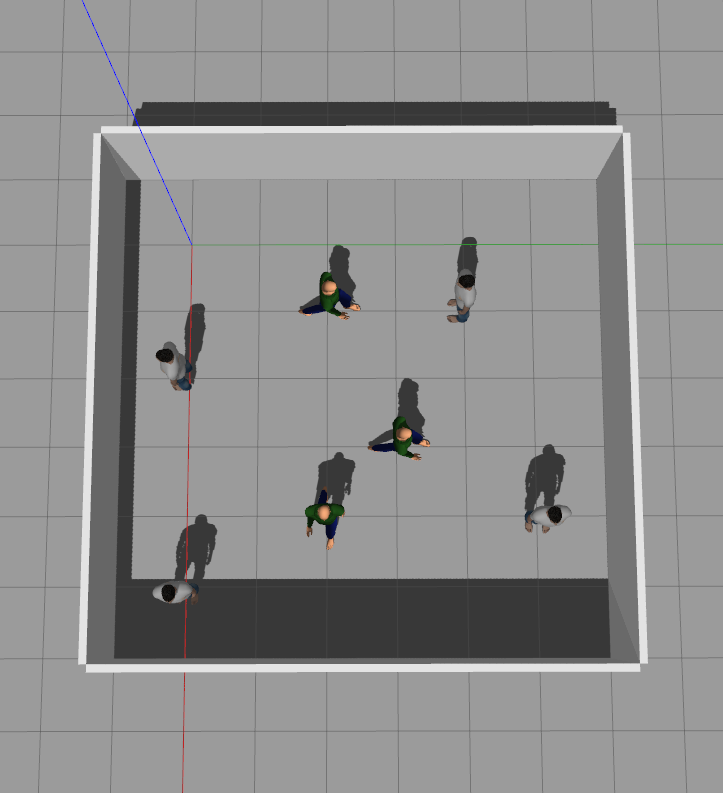}
&
\includegraphics[width=.32\linewidth]{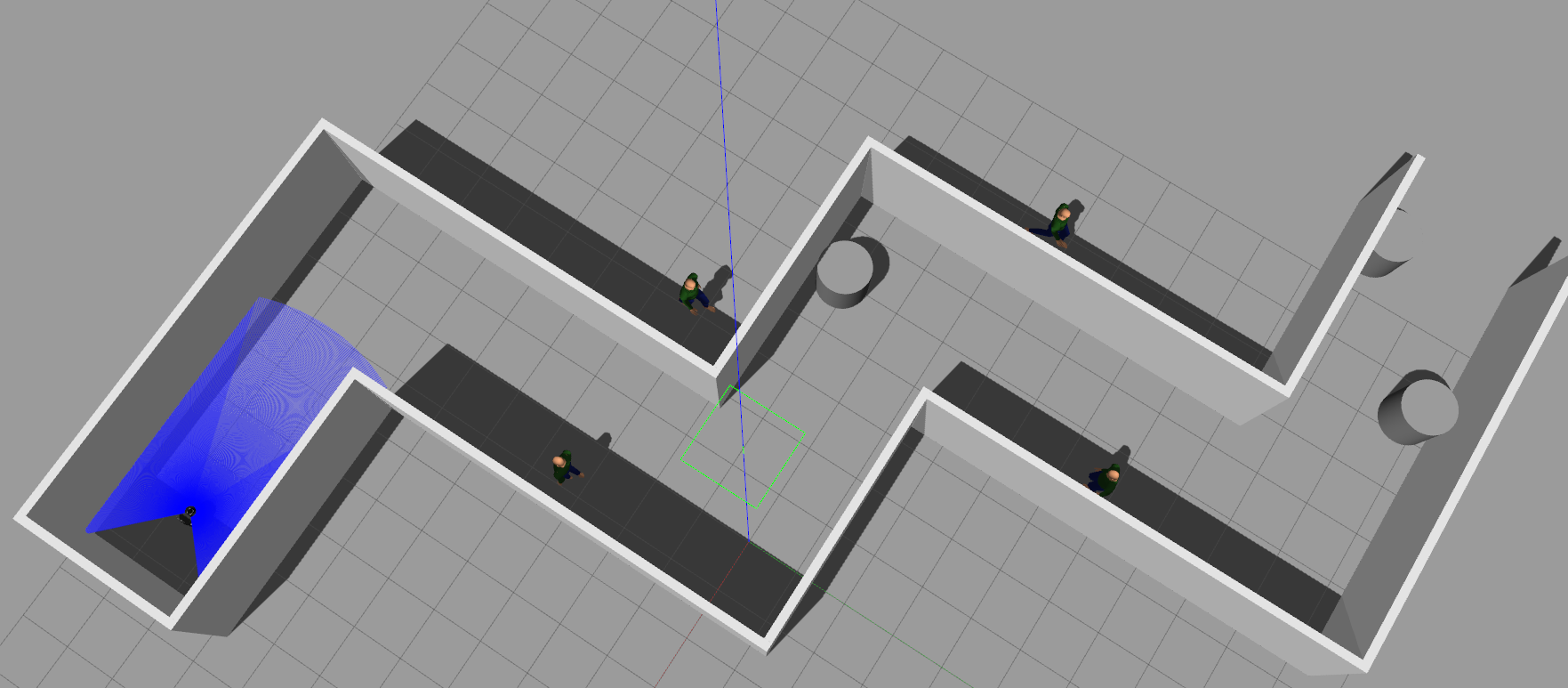}
\end{tabular}
\caption{Left to right: The different training scenarios used by our algorithm from simplest to complex. \textbf{Left}: Static Scenario. \textbf{Middle}: Scenario with static and random dynamic obstacles. \textbf{Right}: Scenario with occlusions and sharp turns. }
\label{fig:TrainingScenarios}
\vspace{-10pt}
\end{figure*}


\textbf{Static Scenario:} 
We use two independent robots that must start from fixed initial locations and head towards fixed goals straight in front of them while avoiding a wall. A single policy is shared between the two robots so that both left and right turning maneuvers are learned simultaneously. Using this initialized policy, we then provide random initial locations and random goals to the two robots for the goal-reaching and static collision avoidance capabilities to generalize well. 

\textbf{Random Static and Dynamic Obstacles:}
In this scenario, a single robot has to move towards a random goal in the presence of static human models and randomly moving pedestrians. We ensure that there is more than enough space through which the robot can navigate. To generate pedestrian trajectories, we assign several waypoints or goals in the environment for the pedestrian models to move to at different time instants. The simulated pedestrians move at near human walking speeds and tend to mimic natural human walking motions. 

\textbf{Scenario with occluded obstacles:}
The saved model from the previous scenario is next trained on a complex corridor scenario, which requires the robot to perform several sharp turns and avoid pedestrians who can only be observed in close proximity. This trains the robot to perform quick maneuvers in real-world scenarios where occluded pedestrians walk towards the robot suddenlyr.

\textbf{Sim-to-real transfer:} Both the scenarios with dynamic obstacles make the sim-to-real transfer and generalization easier, as the depth camera observes noisy 3-D real-life like data such as pedestrian contours and walls, which is fused with more accurate lidar data. Such fusion was not possible in previous methods \cite{JiaPan1, JiaPan2}, as they were trained in a 2.5D simulator. However, in the corridor scenario, there is a danger of the policy overfitting to follow walls and moving to goals which are straight ahead of the robot instead of generalizing to all scenarios. To prevent this, we also parallelly run robots in the simple static scenarios with random goals alongside the corridor scenario. 
\section{Results and Evaluations}
In this section, we describe our implementation and highlight its performance in different scenarios. We also compare our navigation algorithm with prior methods and perform ablation studies to highlight the benefits of implicit multi-sensor fusion and our reward function.


\subsection{Implementation}
We train our model in simulations that were created using ROS Kinetic and Gazebo 8.6 on a workstation with an Intel Xeon 3.6GHz processor and an Nvidia GeForce RTX 2080Ti GPU. We use Tensorflow, Keras and Tensorlayer for implementing our network. 
We use models of the Hokuyo 2-D lidar and the Orbbec Astra depth camera  in Gazebo to simulate sensor data during training and evaluation. The Hokuyo lidar has a range of $4$ meters, FOV of 240$^\circ$ and provides $512$ range values per scan. The Astra camera has a minimum and maximum sensing range of $1.4$ meters and $5$ meters respectively. We use images of size $150 \times 120$ with added Gaussian noise $\mathcal{N}(0,0.2)$ as inputs to our network training. We mount these sensors on a Turtlebot 2 and a Clearpath Jackal robot to test our model in real-world scenarios such as crowded corridors and occluded scenes.


\begin{figure}[t]
\includegraphics[width=\columnwidth,height=7cm]{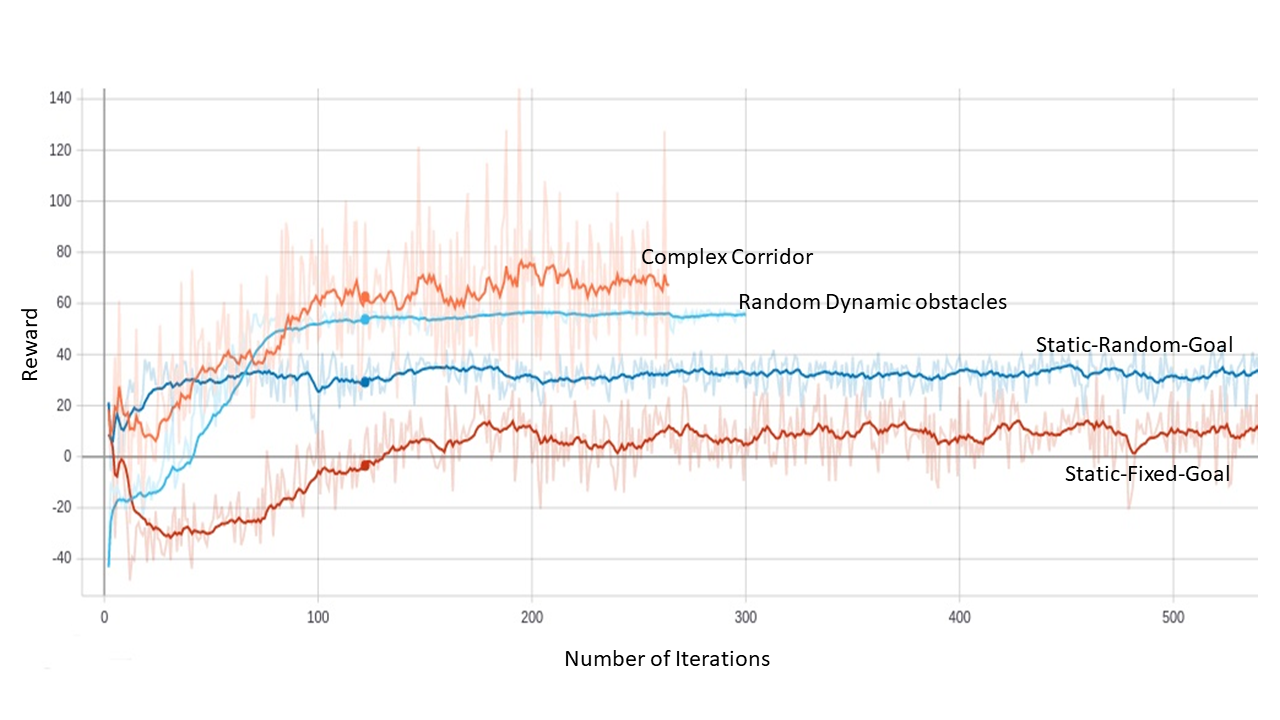}
\caption{Convergence of our reward function vs the number of iterations for different training scenarios. Training in all scenarios converges within $200$ iterations.}
\label{fig:reward}
\vspace{-10pt}
\end{figure}

\subsection{Training Convergence and Data Efficiency}
The convergence of our reward function versus the number of iterations for different training scenarios is shown in Fig.\ref{fig:reward}. The training in all scenarios starts to converge around 100 iterations, and stabilizes around 200 iterations and the total process completes in six days. This increase in training time when compared with previous methods is due to the high dimensionality of the 3-D depth data used during training. Previous methods do not perform any kind of implicit sensor fusion to detect and avoid obstacles. However, the training time does not affect our run-time performance, which can be observed from our real-world tests. 

\begin{table*}[h]
\resizebox{0.8\textwidth}{!}{
\begin{tabular}{|c|c|c|c|c|c|} 
\hline
Metrics \Tstrut & Method & Narrow-Static & Narrow-Ped & Occluded-Ped & Dense-Ped \\ [0.5ex] 
\hline
\multirow{4}{*}{Success Rate} & DWA & 1 & 0.1 & 0.5 & \gray{0.0} \\
 & Depth Camera  & 0.85 & 0.55 & 0.2 & 0.2 \\
 & Fan et al. & 1 & \gray{0.0} & \gray{0.0} & \gray{0.0} \\
 & CrowdSteer & 1 & 1 & 0.9 & 0.67 \\
\hline
\multirow{4}{*}{Avg Trajectory Length} & DWA & 6.33 & 14.86 & 27.2 & \gray{7.46} \\
 & Depth Camera  & 6.18 & 16.30 & 25.7 & 14.95 \\
 & Fan et al. & 6.86 & \gray{6.16} & \gray{13.63} & \gray{9.17} \\
 & CrowdSteer & 6.44 & 15.51 & 27.18 & 16.58 \\
\hline
\multirow{4}{*}{Mean Time} & DWA & 20.9 & 44.1 & 60.4 & \gray{25.72} \\
 & Depth Camera  & 58.7 & 43.6 & 78.20 & 90.73 \\
 & Fan et al. & 106.93 & \gray{13.76} & \gray{28.025} & \gray{38.05} \\
 & CrowdSteer & 34.04 & 41.48 & 70.54 & 64.9 \\
\hline
\multirow{4}{*}{Avg Velocity} & DWA & 0.30 & 0.34 & 0.45 & \gray{0.29} \\
 & Depth Camera  & 0.11 & 0.37 & 0.33 & 0.16 \\
 & Fan et al.
 & 0.06 & \gray{0.44} & \gray{0.48} & \gray{0.23} \\
 & CrowdSteer & 0.20 & 0.37 & 0.39 & 0.26 \\
\hline
\end{tabular}
}
\caption{\label{tab:results} We compare the relative performance of CrowdSteer that uses multiple sensors (depth camera + lidar) with other learning methods that use a single sensor, and a traditional method (DWA) in challenging scenarios. The values in gray are until a collision or oscillation occurred. These numbers clearly highlight the benefit of our novel deep reinforcement learning algorithm (CrowdSteer) that uses multiple sensors over prior methods.}
\end{table*}

\subsection{Testing scenario}
We consider five different test scenarios that have narrower or different sections, as compared to the synthetic datasets used during our training phase. This demands tight maneuvers from the robot to reach the goal. These testing scenarios are more challenging than existing simulation benchmarks and help to better test CrowdSteer's sim-to-real, and generalization capabilities. We define \textit{Least Passage Space} (LPS) as the minimum space available to the robot in-between obstacles when moving towards the goal. The scenarios we consider are:

1. \textbf{Narrow-Static}: Scenario with only static obstacles where the LPS is $<$ 0.7 meters.

2. \textbf{Narrow-ped}: Scenario with 8 pedestrians walking in the opposite direction to the robot's motion in a narrow corridor, with an LPS of $<$ 1.5 meters. 

3. \textbf{Occluded-ped}: Scenario with sharp turns with static obstacles and pedestrians that are occluded by walls. The LPS is $<$ 1 meter.  

4. \textbf{Dense-Ped}: Scenario with 18 pedestrians in a corridor of width 6 meters. Pedestrians may walk together as pairs, which require a robot to make sharp turns in the presence of multiple dynamic obstacles. The LPS is $<$ 1 meter.

\begin{figure}[t]
\includegraphics[width=0.7\columnwidth,height=4cm]{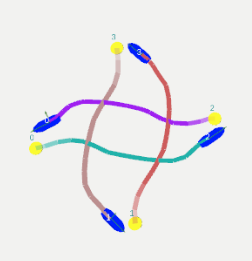}
\caption{Testing the generalization of our training in a scenario with four agents (in blue) moving towards antipodal points on a circle (in yellow). Each robot takes full responsibility for collision avoidance, which leads to making wide maneuvers during collision avoidance.}
\label{fig:Traj_4}
\vspace{-10pt}
\end{figure}

5. \textbf{Circle}: To test the generalization of our method for multi-robot collision avoidance, we make 4 robots move towards antipodal positions on a circle. The trajectories of the 4 robots is shown in Fig. \ref{fig:Traj_4}.

\subsection{Performance Benchmarks and Metrics}
We compare the benefits of our hybrid sensing method with three prior algorithms: (i) DWA~\cite{DWA} which uses a lidar / proximity sensors to detect obstacles along with a global planner which requires a map of the environment.(ii) An implementation that uses a single depth camera that was trained using PPO and our reward functions; (iii) Fan et al.~\cite{JiaPan1}, the current state-of-the-art learning-based collision avoidance method for dense crowd navigation, which uses a single 2-D lidar for sensing. Its real-world implementation is shown in \cite{CrowdMove}. Fan et al.~\cite{JiaPan1} assumes that the pedestrians in dense scenarios would cooperate with the robot to avoid collisions. On the other hand, we assume that the robot takes  the full responsibility to avoid collisions, though the pedestrians may or may not be cooperative.
We use the following  metrics to evaluate the performance of different navigation algorithms:
\begin{itemize}
\item \textbf{Success Rate} - The number of times that the robot reached its goal without colliding with an obstacle over the total number of attempts.

\item \textbf{Average Trajectory Length} - The trajectory length traversed by the robot until the goal is reached, calculated as the sum of linear movement segments over small time intervals over the total number of attempts. In cases where the robot never reached the goal, we report the trajectory length \textit{until it collided or started oscillating} indefinitely.

\item \textbf{Mean Time} - Average time taken to reach the goal over all attempts. If the goal is never reached in all attempts, we report the mean time until a collision or oscillation. 

\item \textbf{Average velocity} - The average velocity of the robot until a collision occurs or the goal is reached over all attempts. 
\end{itemize}

\subsection{Analysis and Comparison}
The results of our experiments and our ablation study to check the benefits of using multiple sensors (both lidar and depth camera) versus using one sensor (only depth camera) are shown in Table \ref{tab:results}.  \\
\textbf{Comparison with DWA:} DWA~\cite{FoxThurn} and CrowdSteer succeed in reaching the goal 100\% of the times in static scenarios. However, as the number and density of dynamic agents in the environment increase, DWA's success rate drops considerably when compared to CrowdSteer. This is due to the re-planning time it takes in the presence of dynamic obstacles or pedestrian. DWA performs well in the occluded-ped scenario due to its use of global map, which eliminates many occlusions. In cases where DWA reached the goal, it has more optimal trajectories and mean time due to the optimal global planner. CrowdSteer manages to have comparable performances without any global knowledge or planner. 

\textbf{Comparison with Fan et al.~\cite{JiaPan1}:} The robot that uses Fan et al.'s method works well in the static scenario. However, while it manages to avoid collisions, it either gets stuck or starts oscillating indefinitely in scenarios with non-cooperative dynamic obstacles and occluded spaces. The values for the trajectory length, mean time and velocity reported when Fan et al's method always failed, are \textit{until a collision or oscillation} to give a sense of how much the robot traversed towards the goal before failing. These results are due to Fan et al.'s training, which is based on multi-agent collision avoidance where all the agents share the responsibility of avoiding collisions. This assumption may not always hold in dense crowds. CrowdSteer therefore manages to outperform Fan et al's method in all scenarios.

Apart from having a much better success rate, CrowdSteer has similar performance as prior methods in terms of average velocities, better trajectory lengths and mean time to Fan et al. which in turn had better success rates, 41.6\% lower time taken to reach the goal, and 14\% higher average speeds than works like NH-ORCA \cite{NH-ORCA}. Therefore, our method outperforms the current state-of-the-art in dense and occluded scenarios.

\begin{figure}[t]
\includegraphics[width=\columnwidth, height=5cm]{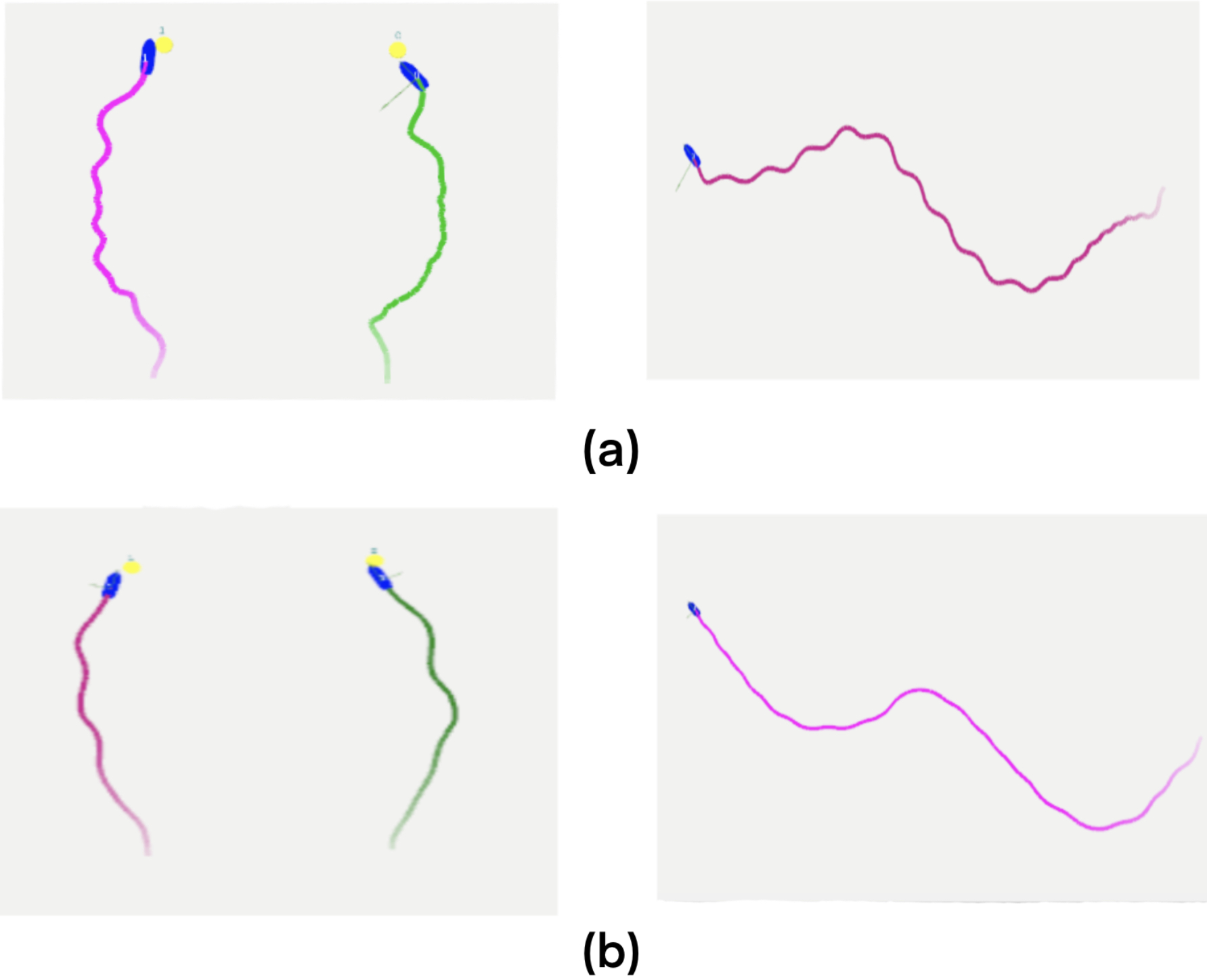}
\caption{\textbf{a)} Trajectories generated by Fan et al.'s method in the Narrow-static (left) and Occluded-ped scenarios (right). \textbf{b)} The trajectories generated by CrowdSteer for the same scenarios.}
\label{fig:traj}
\vspace{-10pt}
\end{figure}

\textbf{Comparing Smoothness:} We also compare the smoothness of the trajectories computed by CrowdSteer and Fan et al.~\cite{JiaPan1} method (Figure \ref{fig:traj}) in the  Occluded-Ped, and Narrow-static scenarios. CrowdSteer is trained to maintain a safe distance from obstacles and avoids oscillations. This results in significant improvement in the smoothness of the trajectories.  

\textbf{Ablation study for multi-sensor fusion:} We compare the effects of using fused data from two sensors (CrowdSteer) versus a model which uses one sensor (Depth Camera) trained with our training scenarios and reward function in different scenarios. Using only a depth camera limits the robot's sensing range and field of view. Due to this, the depth camera model does not have a 100\% success rate in any scenario. This also reflects in the success rate drop in the Occluded-Ped and Dense-Ped benchmarks. However, it still manages to succeed in the dynamic testing scenarios due to our training scenarios and network architecture. Our CrowdSteer algorithm
takes advantage of both the high accuracy of the lidar and the complex features extracted from the depth images and demonstrates significantly better success rates, lower mean time, and higher average velocities.

\textbf{Ablation study of smoothness:} We study the effect of our reward function on the robot's trajectory smoothness. We trained two policies, one including the penalty for oscillations in the reward function (Eqn. \ref{oscillation}), and the other without it. The average number of oscillations in the robot's trajectory are summarized in Table \ref{tab:ablations}. The models are evaluated in two scenarios: (i) Scenario without any obstacles, and the robot moves in a straight line for 10 meters towards the goal, (ii) Scenario where the robot must maneuver to avoid a static obstacle before reaching the goal. Empty/sparse scenarios are used so that turns during dynamic obstacle avoidance does not affect the number of oscillations. We observe that there is a significant reduction in the number of oscillations when the penalty is included. 

\begin{table}[!htb]
  \centering
\resizebox{\columnwidth}{!}{%
    \begin{tabular}{|l|c|c|}
    \hline
    Scenario & Without Penalty & With Penalty \\
    \hline
    Empty world & 9.8 & 2 \\
    \hline
    Static obstacle & 9 & 2 \\
    \hline
  \end{tabular}
}
\caption{The average number of oscillations in two scenarios for two models trained without and with the oscillations penalty term. We see a significant reduction in the number of oscillations in our model that is trained with the penalty.}
\vspace{-15pt}
  \label{tab:ablations}
\end{table}

\textbf{Real-world scenarios:} We use CrowdSteer to navigate a Turtlebot 2 and a Clearpath Jackal robot in crowds with varying densities (~1-3 person/$m^2$), as shown in the video. The robots face high randomness in terms of the direction and velocities of pedestrians, which was not encountered during training. We compare the motion of CrowdSteer with Fan et al.\cite{JiaPan1} method in similar scenarios. Compared to Fan et al., we observe that CrowdSteer has smoother trajectories in both robots and avoids all collisions with the obstacles. On the other hand, Fan et al.\cite{JiaPan1} method to compute collision-free velocities are highly oscillatory. In occluded spaces such as corridors, our CrowdSteer algorithm was able to avoid sudden obstacles which appear in places such as T and L junctions. These tests show the advantages of implicit sensor fusion and our training scenarios with occlusions. Our real-world tests also demonstrate our method's strong sim-to-real and generalization capabilities. 

\textbf{Failure Cases:} CrowdSteer may not work well certain cases. In highly spacious regions, CrowdSteer could exhibit oscillatory behaviors. It might also fail for acute angled turns and in environments with reflective or transparent surfaces, and high interference from infrared light in the surroundings. In crowds with density > 4 people/$m^2$ or scenarios with very minimal or narrow space for navigation, the robot may not find a collision-free path. 

\section{Conclusion, Limitations and Future Work}
We present a novel sensor-based navigation algorithm, CrowdSteer, that simultaneously uses multiple sensors such as 2-D lidars and cameras. Our approach is designed for dense scenarios with pedestrians and makes no assumption about their motion. In practice, our approach works well in complex, occluded scenarios and results in smoother trajectories. Our approach has some limitations and failure cases. It is susceptible to freezing robot problem in very dense settings and the computed trajectories are not globally optimal. Furthermore, the current sensors may not accurately handle glass or non-planar surfaces. There are many avenues for future work. In addition to overcoming these limitations, we need to evaluate its performance in other scenarios and outdoor settings. We may also take into account the dynamics constraints of the robots in terms of trajectory computation.
 

\bibliographystyle{ACM-Reference-Format}  
\bibliography{References}  

\end{document}